\pdfoutput=1

\documentclass[11pt]{article}

\usepackage{acl}

\usepackage{times}
\usepackage{latexsym}
\usepackage{url}
\usepackage[T1]{fontenc}

\usepackage[utf8]{inputenc}

\usepackage{microtype}

\usepackage{inconsolata}

\usepackage{hyperref}       
\usepackage{url}            
\usepackage{booktabs}       
\usepackage{amsfonts}       
\usepackage{nicefrac}       
\usepackage{microtype}      
\usepackage{lipsum}
\usepackage{fancyhdr}       
\usepackage{graphicx}       
\usepackage{float}
\usepackage{xcolor}
\usepackage{algorithm}
\usepackage[noend]{algpseudocode}
\usepackage{stfloats}
\usepackage{multirow}
\usepackage{bbold}
\usepackage{amsmath}
\usepackage{url}
\usepackage{subcaption}
\usepackage{pifont} 
\usepackage{makecell}
\newcommand{\RN}[1]{%
	\textup{\lowercase\expandafter{\it \romannumeral#1}}%
}
\newcommand{\specialcell}[2][c]{%
  \begin{tabular}[#1]{@{}c@{}}#2\end{tabular}}

\title{Self-Checker: Plug-and-Play Modules for Fact-Checking with\\Large Language Models}

\author{\stepcounter{footnote}Miaoran Li\thanks{\ \ This work was done during an internship at Microsoft Research.} \\
  Iowa State University \\
  \texttt{limr@iastate.edu} \\\And
  Baolin Peng \thanks{\ \ Currently at Tencent AI Lab. Work done at Microsoft Research.}\\
  Microsoft Research \\
  \texttt{baolin.peng@microsoft.com} \\\And
  Michel Galley \\
  Microsoft Research \\
  \texttt{mgalley@microsoft.com} \\\AND
  Jianfeng Gao \\
  Microsoft Research \\
  \texttt{jfgao@microsoft.com} \\\And
  Zhu Zhang \\
  University of Rhode Island \\
  \texttt{zhuzhang@uri.edu}
  }

\begin{document}
\maketitle

\def\framework{\textsc{Self-Checker}}
\def\dataset{\textsc{BingCheck}}
\def\policyagent{policy agent}
\def\claimprocessor{claim processor}
\def\querygenerator{query generator}
\def\evidenceseeker{evidence seeker}
\def\classifier{verdict counselor}

\newcommand{\cmark}{\text{\ding{51}}}
\newcommand{\xmark}{\text{\ding{55}}}

\begin{abstract}
Fact-checking is an essential task in NLP that is commonly utilized to validate the factual accuracy of a piece of text. 
Previous approaches mainly involve the resource-intensive process of fine-tuning pre-trained language models on specific datasets. In addition, there is a notable gap in datasets that focus on fact-checking texts generated by large language models (LLMs).
In this paper, we introduce \framework{}, a plug-and-play framework that harnesses LLMs for efficient and rapid fact-checking in a few-shot manner.
We also present the \dataset{} dataset, specifically designed for fact-checking texts generated by LLMs.
Empirical results demonstrate the potential of \framework{} in the use of LLMs for fact-checking. Compared to state-of-the-art fine-tuned models, there is still significant room for improvement, indicating that adopting LLMs could be a promising direction for future fact-checking research.

\end{abstract}

\section{Introduction}
\begin{figure}[!htb]
  \centering  
  \includegraphics[width=\columnwidth]{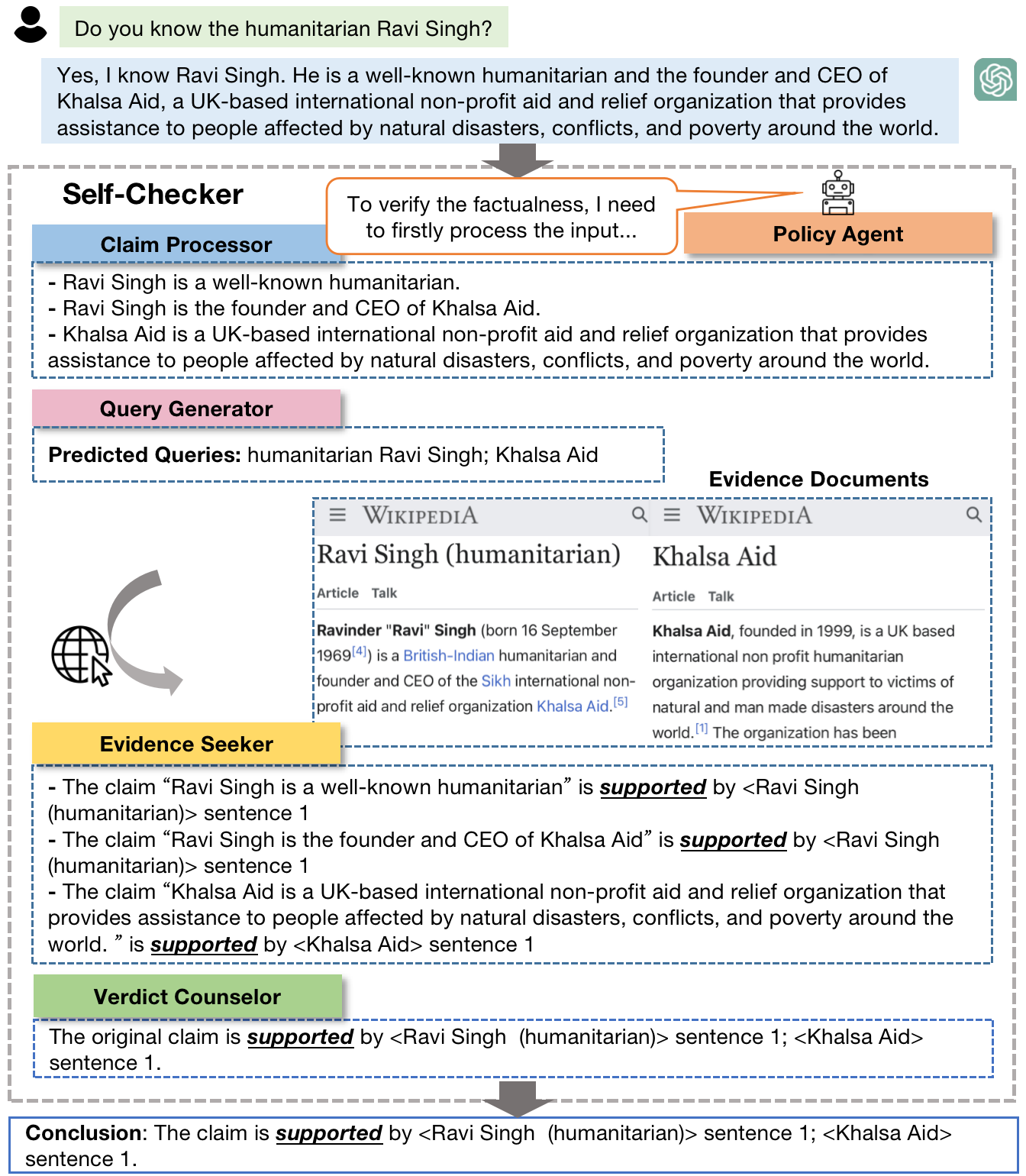}
  \caption{\framework{} assesses the veracity of LLM generated text by (1) extracting simple claims for verification from the input text, (2) generating search queries for retrieval, (3) selecting evidence sentences, and (4) predicting the final conclusion.}
  \label{fig:intro_eg}
\end{figure}

Fact-checking is an essential task in natural language processing, focusing on evaluating the accuracy of text. The advent of large language models (LLMs), such as ChatGPT, GPT-4~\cite{OpenAI2023GPT4TR}, and GPT-3~\cite{brown2020language}, has intensified the importance of this task. As LLMs gain widespread use, the risk of generating false information and hallucinating facts becomes a prominent concern. Despite the extensive implicit knowledge in LLMs and their superior ability to generate realistic responses, ensuring the accuracy and truthfulness of their outputs remains a significant challenge.

Researchers have developed methods for fact-checking and subtasks, including claim detection and fact verification~\cite{guo2022survey}.
Traditional fact-checking approaches typically involve fine-tuning LLMs on specific datasets, which can be computationally expensive and time-consuming. The accelerated progress of LLMs has sparked recent exploration into their potential for fact-checking. \citet{pan-etal-2023-fact} proposed ProgramFC which prompts CodeX for reasoning program generation to guide the verification process.

Existing fact-verification datasets~\cite{thorne2018fever, schuster2021get,petroni2022improving, kamoi2023wice} mainly center on verifying claims from Wikipedia, which do not capture the complexity of lengthy and informative texts generated by LLMs. The lack of a suitable fact-checking dataset tailored for LLM generation poses a challenge in designing and evaluating frameworks in the evolving landscape of LLMs.

In this paper, we introduce \framework{} (depicted in Figure~\ref{fig:intro_eg}), a framework comprising plug-and-play LLM modules for automated fact-checking. The primary objective of \framework{} is to assess the veracity of complex texts (\textit{e.g.}, the response generated by ChatGPT). To achieve this goal, \framework{} first extracts several simple claims for verification from the input and then predicts search queries for these claims to retrieve documents from a knowledge source (\textit{e.g.}, Wikipedia in this example). After obtaining relevant documents, \framework{} selects evidence sentences for each claim from the documents and finally returns a veracity prediction (\textit{e.g.}, whether the original claim is supported by evidence).
We also construct \dataset{} dataset, which focuses on verifying the factual accuracy of texts generated by LLMs. We collect interactions between a simulated user and an LLM and hire human annotators to determine the factualness of LLM's responses. 

This paper makes the following contributions: 
$(\RN{1})$~We introduce \framework{} to utilize LLMs for automatic fact-checking. 
$(\RN{2})$~We construct \dataset{} dataset, which facilitates future research on fact-checking in a more realistic setting.
$(\RN{3})$~We evaluate the effectiveness of \framework{} on the \dataset{} dataset and two fact verification datasets. Our experiments show that \framework{} is capable of generating reasonable results and exhibits considerable potential in the field of fact-checking.
While \framework{}'s performance remains below that of state-of-the-art (SOTA) models for fact verification, our approach does not require any fine-tuning and can be applied to any off-the-shelf LLM.

\section{\framework{} Framework}
\begin{figure}
  \centering  
    \includegraphics[width=\columnwidth]{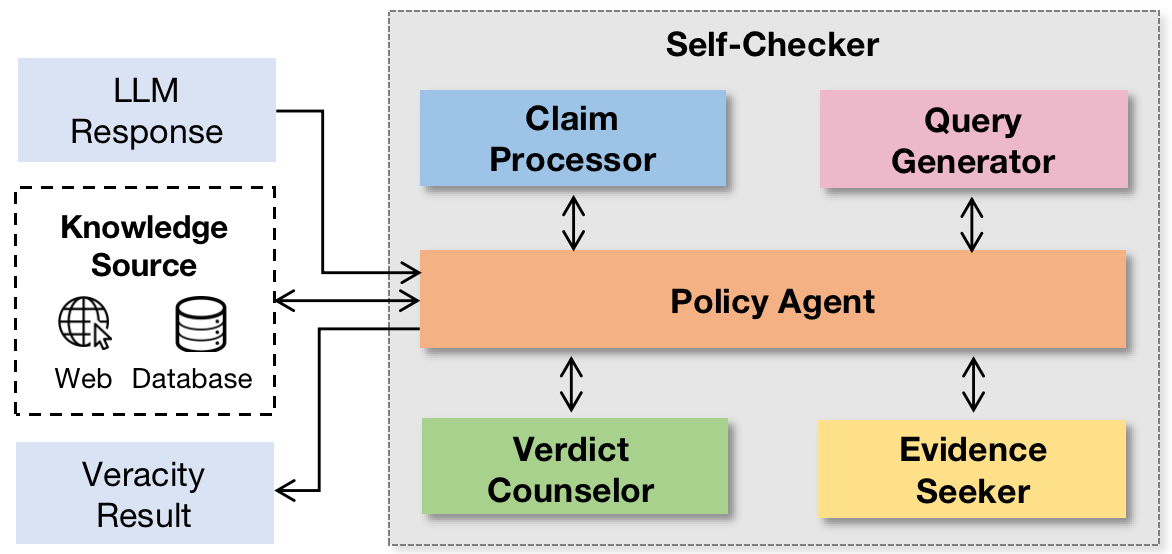}
  \caption{Overview of \framework{}. The framework consists of four plug-and-play modules: (1)~\claimprocessor{}, (2)~\querygenerator{}, (3)~\evidenceseeker{}, and (4)~\classifier{}.}
  \label{fig:framework}
\end{figure}
\framework{} is a framework for fact-checking that is training-free and contains a set of plug-and-play modules---\claimprocessor{}, \querygenerator{}, \evidenceseeker{}, and \classifier{}. The illustration of \framework{} is depicted in Figure~\ref{fig:framework}.
 A comparison of \framework{} against other related frameworks is provided in Table~\ref{tab:model_comparison}. \framework{} is designed to assess the factuality of textual inputs and employs a policy agent that strategically plans future actions based on a predefined set of choices.
 Each module is implemented by prompting an LLM through carefully crafted prompts. Detailed example prompts are provided in Appendix~\ref{sec:prompt}. This modular approach allows for seamless integration to specific fact-checking requirements but also promotes adaptability in diverse application scenarios.

\begin{table*}
\centering
  \resizebox{\textwidth}{!}{
    \begin{tabular}{|c|c|c|c|c|c|}
    \hline
    Method & Goal & Input & \specialcell{Planning\\in the process} & \specialcell{Knowledge\\source} & Output \\
    \hline
    \specialcell{Verify-and-Edit\\\cite{zhao-etal-2023-verify}} & Improve reasoning & CoT reasoning & No & \specialcell{DrQA, Wikipedia,\\Google search} & Revised reasoning\\
    \hline
    \specialcell{FactTool\\\cite{chern2023factool}} & Evaluate factuality & LLM response & No & \specialcell{Wikipedia, Python,\\Calculator, Google scholar} & Factuality labels \\
    \hline
    \specialcell{FActScore\\\cite{min-etal-2023-factscore}} & Evaluate factuality & LLM response & No & Wikipedia & Factuality score \\
    \hline
    \specialcell{FactCheck-GPT\\\cite{wang2023factcheckgpt}} & Correct factual errors & LLM response & No & Google search & Revised response \\
    \hline
    \specialcell{Chain-of-Verification\\\cite{dhuliawala2023chainofverification}} & Correct factual errors & LLM response & Generate entire plan & Parametric knowledge & Revised response \\
    \hline
    \framework{} (Ours) & Evaluate factuality & LLM response & \specialcell{Generate plan\\step by step} & Bing search & Factuality labels \\
    \hline
    \end{tabular}}
    \caption{Comparison of related frameworks. \framework{} aims to provide a factual evaluation of input text, in contrast to FactCheck-GPT and Chain-of-Verification (CoVe), which focus on amending factual inaccuracies in the input text. CoVe revises the input by answering a set of generated verification questions and does not explicitly assess the factuality of the input. While FacTool and FActScore also deliver factual assessment results and FactCheck-GPT can provide intermediate detection results, \framework{} is distinct in that it utilizes a policy agent to dynamically plan future actions from an array of predetermined options.}
    \label{tab:model_comparison}
\end{table*}

\paragraph{Policy Agent}
This module determines the subsequent action of the system from a set of predefined actions. These actions include: 
(1)~calling the \claimprocessor{} to process the complex input,
(2)~requesting search queries from the \querygenerator{},
(3)~retrieving relevant passages from a knowledge source based on the generated search queries,
(4)~utilizing the \evidenceseeker{} to extract evidence sentences for a claim from the retrieved passages,
(5)~requesting the \classifier{} to provide a verdict prediction based on the gathered evidence, and
(6)~sending the final conclusion to the users.

The \policyagent{} follows the task instruction and learns from in-context examples to select the most appropriate action based on the current state and observations of the framework. 
The task description includes a comprehensive list of all available modules, along with brief descriptions of their respective functions. In-context examples provide complete processes of fact-checking for sample input text.
This decision-making process ensures the efficient execution of the fact-checking process. 

\paragraph{Claim Processor} 
The first step in fact-checking is to identify claims for verification from the input text. 
Traditionally, this task involves classifying whether a sentence constitutes a claim or ranking sentences according to their 
check-worthiness~\cite{atanasova2018overview,barron2020overview,zeng_automated_2021}. Leveraging the advanced text generation capabilities of LLMs, we redefine the task of obtaining a set of claims to verify as a generation task.
Given a text $t$ as input, the \claimprocessor{} generates a set of claims $\{c_1, c_2, ..., c_m\}$ that are included in $t$ and need to be verified. If a specific claim for verification has been provided, the \claimprocessor{} can also break it down into a set of simpler claims. Each claim within the set contains a single piece of information, which eases the burden of the subsequent verification process. All generated claims should convey the same information that needs to be verified, as conveyed by the original input.
To achieve this generation process, an LLM is prompted with a combination of task instructions, in-context examples, and a piece of text to be examined. 

\paragraph{Query Generator} 
In order to verify a claim, it is essential to retrieve pertinent information from an external knowledge source. Given a claim $c$, the \querygenerator{} predicts search queries $q = \{q_1, q_2,...,q_k\}$ for the purpose of information retrieval. These generated queries are then used to obtain relevant passages $\{p_1,p_2,...,p_k\}$ from a knowledge source.
The query generation process is accomplished by prompting an LLM. The prompt for the \querygenerator{} includes task instructions, in-context examples, and the claim to be verified.

\paragraph{Evidence Seeker} 
The \evidenceseeker{} aims to identify evidence sentences for a given claim from the retrieved passages. Given a claim $c$ and the set of retrieved passages $\{p_1, p_2, ..., p_k\}$, the \evidenceseeker{} returns a set of selected sentences $\{s_1, s_2, ..., s_n\}$ that indicate the veracity of the claim.
To accomplish this process, an LLM is prompted through a specific prompt comprised of task instruction, in-context examples, the claim to be verified, and the retrieved passages.

\paragraph{Verdict Counselor} 
The primary objective of the \classifier{} is to analyze the set of claims that require verification, together with the corresponding evidence sentences for each claim. This module is responsible for predicting the veracity $r$ of the entire set of claims. By examining the provided evidence, the \classifier{} determines the factuality of each claim and assigns an appropriate veracity label, such as \emph{supported}, \emph{partially supported}, or \emph{refuted}. The labels are then aggregated to obtain the final result of the entire set.
The veracity labels used by the \classifier{} are predefined, encompassing the degrees of entailment (\textit{e.g.}, supported/partially supported/not supported/refuted). To accomplish this process, an LLM is prompted with specific instructions.

\section{The \dataset{} Dataset}
Recent work~\cite{liu2023evaluating} shows that while existing generative search engines powered by LLMs can provide fluent and appear informative responses, they often suffer from hallucination.
To alleviate the problem of hallucinations in LLM generation and facilitate fact-checking research in a more realistic setting, we develop the \dataset{} dataset by human annotation with the assistance of the \framework{} framework. We aim to annotate texts generated by an LLM that are naturally occurring and fine-grained. We collect responses from LLM to user queries related to various topics, which are relatively long and informative. We process complex response into multiple simple claims that are worth-checking and the provide fact-checking information for both response level and claim level. 


\subsection{Dataset Construction}

\subsubsection{Base Data Collection}
To collect responses to various user queries generated by an LLM, we adopt ChatGPT to simulate a curious user and gather responses generated by Bing Chat\footnote{It is named as Bing Chat when we collected the data. It has been updated to Microsoft Copilot now. The implementation is based on \url{https://github.com/acheong08/EdgeGPT}}. We prompt ChatGPT with a user persona characterized by curiosity and an inclination to ask questions on various topics and collect 396 interaction instances between the simulated user and Bing Chat. The responses generated by Bing Chat serve as the input text to be verified. 
\begin{figure*}
  \centering  \includegraphics[width=\textwidth]{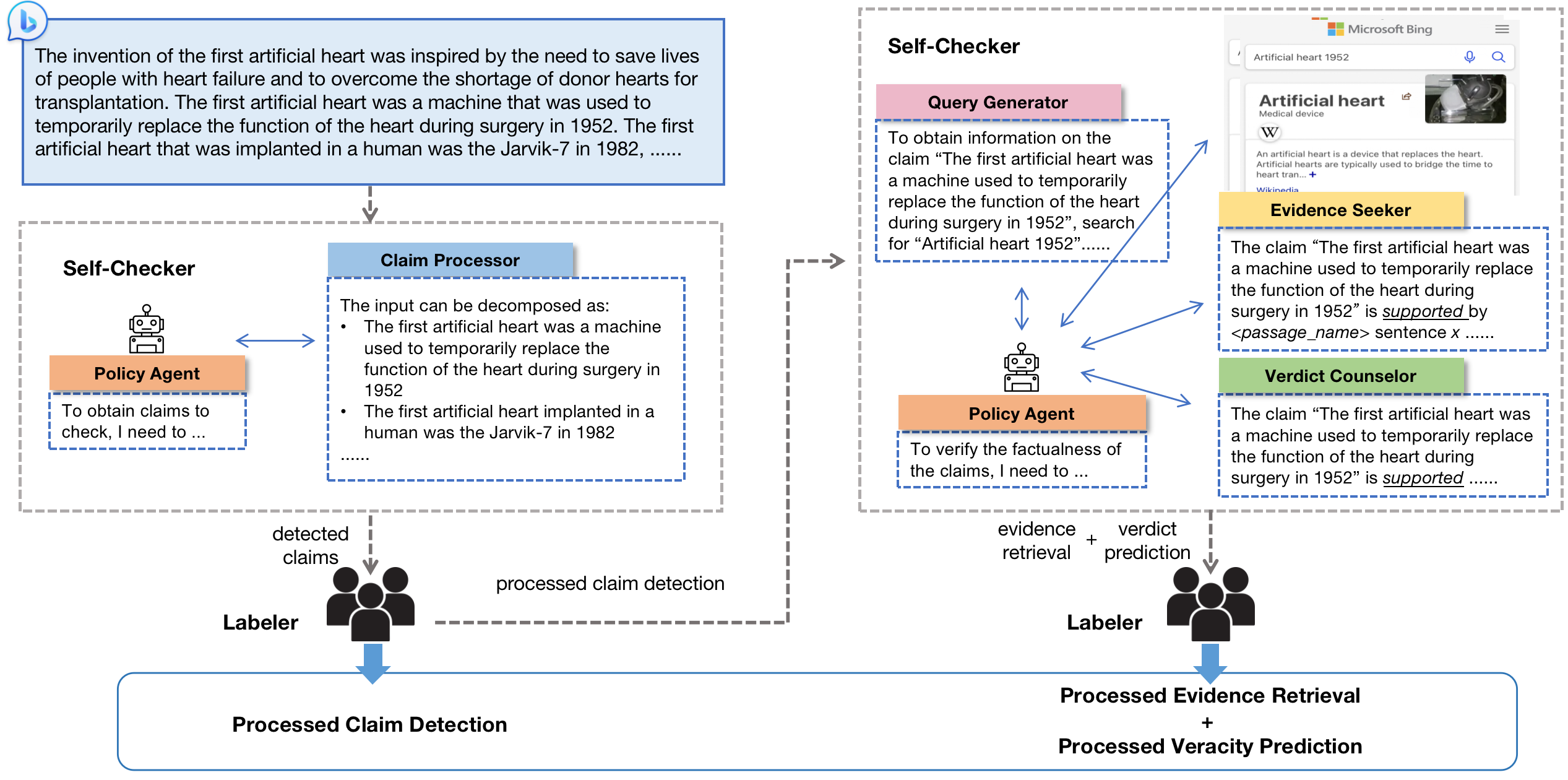}
  \caption{Illustration of \dataset{} dataset construction. The initial claim detection results are obtained using \framework{}, and human annotators verify and refine these automatic results. Processed claims are entered into \framework{} for fact verification data generation, and the outputs are further validated by human workers.}
  \label{fig:dataset}
\end{figure*}

\subsubsection{Data Annotation}
After collecting the base data on the interaction between the simulated user and Bing Chat, we hired human workers on Amazon Mechanical Turk to annotate the data. We aim to autogressively collect annotated data for three subtasks: (1) claim detection, (2) evidence retrieval, and (3) veracity prediction. 
To ensure the quality of data annotation, we have launched onboarding tasks to select proficient workers. Onboarding tasks mirror main tasks but are less demanding. Only qualified workers who pass the onboarding task access the primary task with higher rewards. Each record in \dataset{} is then labeled by a qualified worker. 

Considering the potential challenges and time constraints associated with human annotation, we adopt the \framework{} framework to assist in the following annotation process. The main idea is that for each subtask, we first utilize the \framework{} framework to generate candidate solutions to a subtask and then require human annotators to validate and correct the candidate solutions. The processed solutions are used to generated candidate solutions to the next subtask. The human-processed data are collected in \dataset{}. The data collection process is depicted in Figure~\ref{fig:dataset}. The instruction for human annotation and an example of annotated data are shown in Appendix~\ref{sec:dataset}.

\paragraph{Claim Detection}
Using the \framework{} framework, particularly the \claimprocessor{} module, we generate a set of claims for verification for each input. Human workers then assess and correct the automatically labeled data. Workers receive a Bing Chat response and a set of claims extracted by \framework{}. Their task involves selecting all claims in the response that necessitate verification from the provided set and filling in any missing claims requiring verification but not included in the given set.

\paragraph{Evidence Retrieval and Veracity Prediction}
Claims processed by workers are inputted into the \framework{} framework, integrating the \querygenerator{}, \evidenceseeker{}, and \classifier{} modules. For each claim, \framework{} predicts a search query, retrieves relevant passages from a certain knowledge source,\footnote{In our implementation, we utilized the Bing search engine and retrieved three passages for each claim.}, selects evidence sentences, and predicts the candidate veracity label. 

We consider four veracity labels: \textsc{Supported}, \textsc{Partially Supported}, \textsc{Refuted}, \textsc{Not Supported}. A claim is refuted if any evidence sentence contradicts it. A claim is supported if there are no refuting sentences and at least one sentence supporting it. A claim is partially supported when there are sentences that contribute to the credibility of a portion of the claim but do not fully establish its truth or validity. A claim is not supported if there are no sentences that refute, support, or partially support the claim.

The automatic results of evidence retrieval and claim verification are provided to workers.  Their task involves reviewing the claim along with each automatically selected evidence sentence, selecting all sentences relevant to verifying the claim's factuality. Finally, the workers determine the verdict results based on their selection.

\subsection{Statistics}
\begin{table}
  \centering
  \resizebox{\columnwidth}{!}{
  \begin{tabular}{c|cc}
    \toprule
    Statistic & Response & Extracted Claim\\
    \midrule
    Total number & 396 & 3840\\
    Average length & 391.5 & 26.3\\
    Number of evidence sentences & 55.0 & 6.2\\
    Number of claims per response & 9.7 & -\\
    \bottomrule
  \end{tabular}}
    \caption{Statistics of the \dataset{} dataset. The ``Response'' column stands for raw response generated by BingChat, and ``Extracted Claim'' represents a claim extracted from a response that needs to be verified. The number of evidence sentences is computed only on responses/claims with \textsc{Supported}, \textsc{Partially Supported}, \textsc{Refuted} labels. } 
    \label{tab:statistics}
\end{table}
Table~\ref{tab:statistics} presents the overall statistics for the \dataset{} dataset. The original responses generated by Bing Chat have an average length of 391.5 tokens and can be decomposed into an average of 9.7 claims for verification. The dataset contains more than 3800 claims. For claims that are refuted, supported, or partially supported, there are approximately 6 evidence sentences on average.

Table~\ref{tab:dataset_comparison} presents a comparative analysis of BingCheck against established datasets in the fact-checking field. 
Our dataset is characterized by its considerably longer responses compared to those found in the existing datasets. This significant increase in response length suggests that BingCheck can provide a more complex and extensive framework for assessing factuality. Furthermore, this increased length underscores the alignment of our dataset with real-world scenarios, wherein responses to complex or broad inquiries posed to LLMs are typically extensive and detailed, thereby making the factuality evaluation more challenging.

\begin{table*}
  \centering
  \resizebox{\textwidth}{!}{
    \begin{tabular}{|c|c|c|c|c|c|c|c|}
    \hline
    
    \multirow{2}{*}{Dataset} & \multicolumn{2}{c|}{Input}& \multirow{2}{*}{\specialcell{Claim\\granularity}}& \multirow{2}{*}{Knowledge Source} & \multirow{2}{*}{\specialcell{Evidence\\provided}} & \multirow{2}{*}{Task} & \multirow{2}{*}{Scenario}\\
    \cline{2-3}
    &Length & Generated by & &  &  &  &  \\
    \hline
    \specialcell{Fever\\\cite{thorne2018fever}} & 7.3 &	Human & fact & Wikipedia & Yes & Fact verification & Wikipedia claim\\
    \hline
    \specialcell{WiCE\\\cite{kamoi2023wice}} & 24.2 & Human & Fact & Wikipedia & Yes & Entailment classification & Wikipedia claim\\
    \hline
    \specialcell{FactTool\\\cite{chern2023factool}} & 76.3 & ChatGPT & Fact & \specialcell{Wikipedia,\\Python,\\Calculator,\\Google scholar} & Yes & Fact checking & \specialcell{QA, Code, Math,\\Literature review} \\
    \hline
    \specialcell{HaluEval\\\cite{li-etal-2023-halueval}} & 82.0 & ChatGPT & Response & Parametric knowledge & No & Fact checking & \specialcell{QA, Dialog,\\Summary} \\
    \hline
    \specialcell{FELM\\\cite{chen2023felm}} & 89.1 & ChatGPT & Segment & Google search & Yes & Fact checking & \specialcell{World knowledge,\\Science, Math,\\Recommendation} \\
    \hline
    \specialcell{FActScore\\\cite{min-etal-2023-factscore}} & 154.5 & \specialcell{InstructGPT,\\ ChatGPT,\\ PerplexityAI} & Response & Wikipedia & No & Fact checking & \specialcell{Biography generation,\\Long-form response} \\
    \hline
    \specialcell{FactCheck-GPT\\\cite{wang2023factcheckgpt}} & 95.8 & \specialcell{ChatGPT,\\GPT4} & Response & Google search & Yes & \specialcell{Fact checking and\\error correction} & QA \\
    \hline
    \specialcell{Chain-of-Verification\\\cite{dhuliawala2023chainofverification}} & - & Llama-65B & - & Search engine & - & Factual error correction & QA \\
    \hline
    \dataset{} (Ours) & 391.5 & Bing Chat & Response & Bing search & Yes & Fact checking& \specialcell{Long-form response,\\QA} \\
    \hline
    \end{tabular}
    }
    \caption{Comparison of factuality evaluation datasets. The ``Scenario'' column describes the tasks used to gather the initial responses. The critical point of differentiation for our dataset is the significantly greater average response length, which is considerably longer than those in the datasets we have compared it with. }
    \label{tab:dataset_comparison}
\end{table*}

\subsection{Dataset Quality Evaluation}
To evaluate the quality of the annotated data, we have hired Amazon Mechanical Turk workers to perform annotation review tasks. For each annotated record, we have employed three workers to evaluate it. Each worker answers a series of single-choice questions to assess the quality of the annotation. To evaluate the quality of claim detection, a worker is presented with an original response and an annotated list of claims. The workers need to determine whether all listed claims need verification and whether all claims in the response that require verification are included in the given set of claims. To assess the quality of annotations for evidence retrieval and veracity prediction, a worker is presented with a claim and a list of evidence sentences. A worker first determines whether all evidence sentences are relevant for verifying the claim's factualness. Then the worker determines whether the assigned label is correct. We use a majority vote to aggregate the evaluation results.

In terms of claim detection, among all 396 records, the extracted claims in 381 records are deemed comprehensive and verifiable. However, there are 15 records where the claim detection is either missing or contains claims that do not require verification. Regarding evidence retrieval and veracity prediction, we have a total of 3840 extracted claims. Evaluators have found that 94\% of these claims have appropriate evidence sentences. In the case of the remaining claims, there may be redundant and irrelevant sentences within the selected evidence. For verdict prediction, 96\% claims have been considered to be accurately assigned with appropriate labels based on the annotated evidence. There may be some level of noise in the human evaluation results. Nevertheless, this evaluation process provides an estimation of dataset quality and offers valuable insights for further checks and improvements in data annotation.

\section{Experiments}

\subsection{Datasets} 
We evaluate the performance of the \framework{} framework for the fact-checking task on the \dataset{} dataset. Additionally, we assess its efficiency in performing fact verification using the FEVER dataset~\cite{thorne2018fever} and text entailment using the WiCE dataset~\cite{kamoi2023wice}.

\paragraph{\dataset{} Dataset}
The fact-checking process of LLM response in the \dataset{} dataset involves four subtasks: (1)~Claim detection: Given a long paragraph $t$, models are required to generate a set of claims $\{c_1,c_2,...,c_m\}$ that require evidence or proof to support their accuracy or truthfulness. (2)~Document retrieval: Given a claim $c$, models are expected to predict search queries $\{q_1, q_2, ...,q_k\}$ to retrieve relevant articles from a knowledge source. (3)~Sentence retrieval: Given a claim $c$ and relevant passages $\{p_1, p_2, ...,p_k\}$, models are required to select evidence sentences $\{s_1,...,s_n\}$ from the articles. These evidence sentences can either (partially) support or refute the claim, depending on the veracity label design. (4)~Verdict prediction: Given a claim $c$ and the evidence sentences $\{e_1, ..., e_n\}$, models are required to predict the veracity label. The fact-checking process requires the \claimprocessor{}, \querygenerator{}, \evidenceseeker{}, and \classifier{} modules.

\paragraph{FEVER Dataset}
In the FEVER~\cite{thorne2018fever} dataset, claims consist of a single piece of information and do not require further decomposition. The verification of a claim in FEVER involves document retrieval, sentence retrieval and verdict prediction. The FEVER dataset uses three identification labels: \textsc{Supported}, \textsc{Refuted}, and \textsc{NotEnoughInfo}. A claim is verified as \textsc{NotEnoughInfo} if there is insufficient information in Wikipedia to support or refute the claim, either because the claim is too general or too detailed. The dataset provides the names of evidence Wikipedia passages and the indices of evidence sentences. In the verification process, the names of evidence articles serve as search queries. To verify a claim in the FEVER dataset, the \framework{} framework adopts \querygenerator{}, \evidenceseeker{}, and \classifier. We follow the experiment setting in the previous research~\cite{zhao-etal-2023-verify} and use the same subset of Fever.

\paragraph{WiCE Dataset}
The WiCE dataset is specifically designed for verifying Wikipedia citations and consists of claims grounded in cited articles from Wikipedia. Unlike the FEVER dataset, the claims in WiCE contain multiple pieces of information. The verification process in WiCE involves claim detection, sentence retrieval, and verdict prediction.
Complex claims in WiCE are decomposed into simpler subclaims. Verifying claims in WiCE primarily entails sentence retrieval for the cited articles and subsequent verdict prediction. The veracity labels in WiCE include \textsc{Supported}, \textsc{PartiallySupported}, and \textsc{NotSupported}. A claim is classified as \textsc{PartiallySupported} if some tokens within the claim are not supported by any evidence sentence. The prediction results are collected at subclaim levels. The veracity label of the original claim is set to \textsc{Supported} or \textsc{NotSupported}, depending on whether all subclaims are supported or not supported. Otherwise, the original claim is considered \textsc{PartiallySupported}. 
To verify a claim in the WiCE dataset, the \framework{} framework adopts \claimprocessor{}, \evidenceseeker{}, and \classifier{} modules.

\subsection{Experimental Setup}

\paragraph{Implementation} All modules in the \framework{} are implemented using OpenAI GPT-3.5 (text-davinci-003) API with temperature 0.2. The prompt for \policyagent{} consists of three examples due to the length constraint. The prompts for \claimprocessor{}, \querygenerator{}, \evidenceseeker{}, and \classifier{} contain fifteen examples. As for the knowledge source, we employ Bing search engine for BingCheck and Wikipedia for FEVER. Up to three retrieved passages are considered for further evidence selection. In the implementation, we stored FEVER preprocessed Wikipedia passages in a database. The retrieval mechanism automatically incorporates passages whose titles precisely match the generated search query or exhibit partial alignment with the predicted search query. 

\paragraph{Evaluation Metrics} 
We report label accuracy and F1 score for evidence retrieval, which is computed between all predicted sentences and the golden evidence sentences for claims requiring evidence. Consistent with baseline studies~\cite{kamoi2023wice, thorne2018fever}, we present the F1 score for verdict prediction on the WiCE dataset and the FEVER score for results on the FEVER dataset. The FEVER score is the strict accuracy with the requirement of providing correct evidence for the \textsc{Supported}/\textsc{Refuted} predictions.

\paragraph{Baselines}
We evaluate \framework{} against various methods.
Standard prompting directly predicts verdict labels based on input claims, while Chain-of-thought prompting~\cite{wei2022chain} generates explanations before making predictions.
ReAct~\cite{yao2023react} follows a reason-and-act framework with an external knowledge source\footnote{ReAct is not evaluated on the WiCE dataset as the knowledge retrieval is not included in the verification process for claims in WiCE.}. The setup of the knowledge source is similar to that in \framework{}. 
We also compare with a related method Verify-and-Edit~\cite{zhao-etal-2023-verify} on Fever dataset.
These prompt-based methods are implemented using the OpenAI GPT-3.5 API. 

In addition, we compare our approach to the initial baseline model~\cite{thorne2018fever} and the state-of-the-art (SOTA) model BEVERS~\cite{dehaven2023bevers} on the FEVER dataset.
The baseline model consists of a DrQA~\cite{chen2017reading} document retrieval module, a DrQA-based sentence retrieval module, and an entailment module based on decomposable attention~\cite{parikh2016decomposable}. The SOTA model adopts BERT for evidence retrieval and claim verification, along with meticulous hyperparameter tuning.
For the WiCE dataset, we include the initial baseline model~\cite{kamoi2023wice}, implemented by fine-tuning T5-3B~\cite{raffel2020exploring} on WiCE.

\subsection{Main Results}
\paragraph{Evaluation Results on \dataset{} Dataset} 
The evaluation results on \dataset{} are presented in Table~\ref{tab:bingcheck}. We observe the inherent challenge LLMs face when determining the factualness of complex paragraphs based solely on pre-trained parametric knowledge. It is notable that LLMs prompted with standard and chain-of-thought prompts tend to align with the input, tending to recognize it as supported information. The integration of external knowledge contributes to the improvements in fact-checking. However, a performance gap persists between baseline models and the proposed framework, which underscores the importance of incorporating modules capable of decomposing complex paragraphs into simpler claims, conducting explicit analysis of retrieved passages, and predicting verdicts.
Furthermore, the availability of intermediate results from the fact-checking process enhances our ability to identify performance bottlenecks within \framework{}, making it possible to guide further improvements. 
Despite the introduction of \framework{}, there are limitations in achieving optimal results on \dataset{}, highlighting the inherent difficulty in fact-checking LLM-generated content and prompting further exploration.

\begin{table}
  \centering
  \resizebox{\columnwidth}{!}{
  \begin{tabular}{c|c|ccc}
    \toprule
    \multirow{2}{*}{Model} & \multirow{2}{*}{Accuracy} & \multicolumn{3}{c}{Evidence Retrieval}\\
     & & F1 & Precision & Recall\\
    \midrule
    Standard Prompt & 19.4 & - & - & - \\
    Chain-of-Thought & 15.7 & - & - & - \\
    ReAct~\cite{yao2023react} & 21.0  & - & - & - \\
    \midrule
    \framework{} & 63.4 & 45.0 & 30.5 & 86.1\\
    \bottomrule
  \end{tabular}}
    \caption{Evaluation results on \dataset{}. The accuracy is computed on the response level.}
    \vspace{-0.2cm}
    \label{tab:bingcheck}
\end{table}

\paragraph{Evaluation Results on FEVER Dataset}
\begin{table*}
  \centering
  \resizebox{0.9\textwidth}{!}{
  \begin{tabular}{c|c|c|c|ccc}
    \toprule
    \multirow{2}{*}{Model} & \multirow{2}{*}{Fine-tuning} & \multirow{2}{*}{FEVER Score} & \multirow{2}{*}{Accuracy} & \multicolumn{3}{c}{Evidence Retrieval}\\
    & & & & F1 & Precision & Recall \\
    \midrule
    Standard Prompt & \xmark & - & 49.9 & - & - & - \\
    Chain-of-Thought & \xmark & - & 51.8 & - & - & - \\
    ReAct & \xmark & - & 51.4 & - & - & -\\
    Verify-and-Edit & \xmark & - & 53.9 & - & - & - \\
    \midrule
    \framework{} & \xmark & 47.9 & 56.7 & 47.5 & 75.3 & 34.7\\
    \midrule
    DrQA~\cite{thorne2018fever} & \cmark & 31.9 & 50.9 & 17.5 & 10.8 & 45.9\\
    BEVERS~\cite{dehaven2023bevers} & \cmark & 77.7 & 80.2 & - & - & -\\
    \bottomrule
  \end{tabular}}
    \caption{Evaluation results on FEVER dataset. ``Fine-tuning'' stands for whether the training procedure is required. Verify-an-Edit is experimented with three different knowledge sources~\cite{zhao-etal-2023-verify}. We compare with the highest accuracy obtained by using the Google search engine as a knowledge source.}
    \label{tab:fever}
\end{table*}
The evaluation results on the FEVER dataset are presented in Table~\ref{tab:fever}. Compared to prompt-based baselines, \framework{} improves verification accuracy with explicit evidence retrieval results.
Comparing the performance of the baselines and \framework{}, we observe that LLMs possess a robust capacity to learn from few examples and perform various tasks, including query generation, retrieval and verdict prediction. However, the significant performance gap between the SOTA model and the \framework{} highlights the need to improve the efficiency of the \framework{}.

\paragraph{Evaluation Results on WiCE Dataset}
\begin{table*}
  \centering
  \resizebox{0.78\textwidth}{!}{
  \begin{tabular}{c|c|c|c|ccc}
    \toprule
    \multirow{2}{*}{Model} & \multirow{2}{*}{Fine-tuning} &  \multirow{2}{*}{F1} & \multirow{2}{*}{Accuracy} & \multicolumn{3}{c}{Evidence Retrieval}\\
    & & & & F1 & Precision & Recall \\
    \midrule
    Standard Prompt & \xmark & 9.0 & 65.9 & - & - & - \\
    Chain-of-Thought & \xmark & 36.7 & 50.0 & - & - & - \\
    \midrule
    \framework{} & \xmark & 47.7 & 71.5 & 60.5 & 71.4 & 52.5\\
    \midrule
    T5-3B~\cite{kamoi2023wice} & \cmark & 65.3 & 77.1 & 67.4 & 65.0 & 81.7\\
    \bottomrule
  \end{tabular}}
    \caption{Evaluation results on WiCE test set. ``Fine-tuning'' stands for whether the training procedure is required. Note that we compare with T5-3B model finetuned on WiCE dataset~\cite{kamoi2023wice}.}
    \label{tab:wice}
\end{table*}
The evaluation results for the WiCE dataset are in Table~\ref{tab:wice}. The F1 score for label prediction is quite low for the LLM with standard prompting, as it tends to predict the supported claim as partially or not supported.
In line with earlier findings, \framework{} demonstrates superior efficiency compared to the prompt-based baselines.
A noticeable performance gap emerges when comparing \framework{} with the model fine-tuned on the WiCE dataset. Specifically, in evidence retrieval, \evidenceseeker{} tends to overlook evidence in the passages, highlighting a potential bottleneck in overall performance.


\subsection{Ablation Study}
To assess the impact of each module on overall performance, we conduct an ablation study on three datasets. The evaluation results on \dataset{} are shown in Table~\ref{tab:bingcheck_ab}. 
The first row reflects end-to-end fact-checking performance, encompassing claim detection, document retrieval, sentence retrieval, and verdict prediction. When comparing the first and second rows, we note that providing golden claims results in improvements across all metrics. The marginal difference between results with and without golden documents suggests the low-temperature setting of the API in the \querygenerator{} module ensures stable search query generation, with retrieval results for a fixed query exhibiting consistency.
Even with golden evidence sentences, the label accuracy at the response level does not exceed 70, indicating potential for further enhancements in the \classifier{} module to improve the accuracy of veracity prediction.
In terms of evidence retrieval performance, it is unsurprising to observe an inclination to over-select more evidence sentences. This behavior stems from the dataset construction process, where human workers filter evidence sentences selected by \framework{}, removing less relevant ones.

Analyzing the incorrect predictions with golden evidence sentences, we observe a tendency in \framework{} to be overly optimistic, classifying claims that are only partially supported as fully supported. For instance, the claim ``Brain virus was released on 19 January 1986 by two brothers from Pakistan, Basit and Amjad Farooq Alvi.'' is partially supported by the evidence sentence ``In 1986, Brain was developed by the Pakistani brothers Basit and Amjad Farooq Alvi, who were annoyed at having their heart monitoring software copied for free.'' However, \framework{} overlooks the lack of mention of the exact release date of the Brain virus and predicts the claim as supported based on the evidence.

\begin{table}
  \centering
  \resizebox{\columnwidth}{!}{
  \begin{tabular}{c|cc|c|ccc}
    \toprule
    Golden & \multicolumn{2}{c|}{Golden Evidence} &  \multirow{2}{*}{Accuracy} & \multicolumn{3}{c}{Evidence Retrieval}\\
    Claims&Document & Sentence & & F1 & Precision & Recall\\
    \midrule
    \xmark & \xmark & \xmark & 63.4 & 45.0 & 30.5 & 86.1\\
    \cmark & \xmark & \xmark & 64.3 & 48.8 & 32.7 & 96.5\\
    \cmark & \cmark & \xmark & 64.3 & 49.0 & 32.8 & 97.0\\
    \cmark & \cmark & \cmark & 67.2 & - & - & -\\
    \bottomrule
  \end{tabular}}
    \caption{Ablation results on \dataset{}. ``Golden Claims'' indicates whether the golden claims are given. ``Golden Evidence'' indicates whether the golden documents and sentences are provided.}
    \vspace{-0.2cm}
    \label{tab:bingcheck_ab}
\end{table}

\section{Related Work}
The framework for automated fact-checking involves claim detection and factual verification~\cite{zeng_automated_2021, guo2022survey}. Claim detection identifies statements needing verification, while factual verification includes evidence retrieval and assessment of claim validity.

Claim detection has been approached as a binary classification task, determining if a sentence represents a claim~\cite{hassan2017toward}, or as a ranking task, ordering sentences based on their check-worthiness~\cite{jaradat-etal-2018-claimrank}.

Fact verification requires models to assess the veracity of a given claim by examining evidence information. FEVER dataset~\cite{thorne2018fever} is one of the most popular datasets in this area, and fueled the development of fact verification models~\cite{soleimani2020bert, jiang2021exploring, krishna2022proofver}. The fact verification in FEVER dataset consists of document retrieval, sentence selection, and verdict prediction.

The Vitamin C dataset~\cite{schuster2021get} is proposed for a contrastive fact verification paradigm which requires models to be sensitive to changes in evidence and claims. The WAFER dataset~\cite{petroni2022improving} contains instances from Wikipedia inline citations. The WiCE dataset~\cite{kamoi2023wice} provided fine-grained annotation of supporting evidence and non-supported tokens in claims.

While many work focused on verifying claims against raw text evidence, other recent datasets cover verification against various evidence, such as table~\cite{chen2019tabfact,gupta2020infotabs,akhtar2022pubhealthtab}, knowledge graph~\cite{9576612, vedula2021factkeg, kim-etal-2023-factkg} and other multimodal evidence~\cite{alam2022survey}.

Factual error correction is a task closely related to fact-checking. After assessing the factualness of claims within the input text, a subsequent step is addressing any inaccuracies to improve factual integrity.
Recent studies have explored methods for refining the factualness of text outputs by leveraging retrieved evidence~\cite{thorne-vlachos-2021-evidence, iv-etal-2022-fruit, huang-etal-2023-zero}. In addition to approaches specialized in correcting factual errors, some recent frameworks first assess the factualness of its initial generation and then amend any detected inaccuracies to enhance the overall veracity of the generation~\cite{wang2023factcheckgpt, dhuliawala2023chainofverification, fatahi-bayat-etal-2023-fleek}.

\section{Conclusion and Future Work}
We present \framework{}, a framework for automated fact-checking with plug-and-play modules implemented through prompting LLMs. Additionally, we introduce the \dataset{} dataset, which serves as a valuable resource for future research in fact-checking of LLM-generated responses.
Experimental results demonstrate the significant potential of \framework{} in the fact-checking task.

In future work, a key direction to explore is to enhance the efficiency of \framework{}. One potential avenue is the incorporation of additional working memory to accelerate the verification process by using past information. Furthermore, investigating more efficient strategies for utilizing LLMs in each subtask of fact-checking holds promise for optimizing performance.

\newpage
\section*{Limitations}
One limitation of \framework{} is its inability to account for information updates. If there is out-of-date information that contradicts a claim, \framework{} may classify the claim as refuted even if it is actually supported by the most up-to-date information. This limitation arises due to the mixed and unrefined sources of information used by \framework{} during the fact-checking process. \framework{} does not contain a module to postprocess and filter the retrieved articles.
Another limitation of \framework{} is its high computational cost due to the involvement of multiple chained LLM calls in the process of fact-checking. To ensure the reliability of predictions, we adopt the majority voting approach by running \evidenceseeker{} and \classifier{} multiple times. Although this approach can improve accuracy and stability, it may result in slower response times. However, we anticipate that this limitation can be mitigated in the future with the advancement of more efficient and accessible LLMs. In addition, we will explore providing options to achieve a balance between accuracy and waiting time, allowing users to make informed trade-offs based on their specific requirements.
Another limitation is the sensitivity of \framework{} to prompts. In our preliminary experiments, we have observed variations in performance when using different prompts. Enhancing the robustness of LLMs to prompts is an avenue for future exploration, aiming to improve the reliability and consistency of \framework{}. Furthermore, the current prompts are manually designed, which may be heuristic in nature. We consider investigating automated methods for selecting in-context learning examples and generating strong prompts in the future work.
Additionally, the selection of hyperparameters in \framework{} currently relies on heuristics. Exploring more efficient automated approaches for hyperparameter tuning could improve the overall efficiency of the framework.

A potential limitation of the \dataset{} dataset is the potential bias during annotation. The classification of the veracity of a claim can be subjective. It is important to consider this factor when interpreting and utilizing the \dataset{} dataset for research purposes.

\section*{Ethics Statement}
In this work, we focus on utilizing \framework{} to tackle the problem of hallucinations in the generation results of LLMs. However, it is important to acknowledge that LLMs' generation can also exhibit other potential issues, including the production of offensive and harmful content. Currently, \framework{} does not address these problems. To mitigate these concerns, future work on \framework{} could incorporate a dedicated module specifically designed to detect and remove offensive and harmful content.

\bibliography{references,custom}

\clearpage
\appendix
\section{Example Prompts for \framework{}}
\label{sec:prompt}
The example prompts for modules in \framework{} are shown in Figure~\ref{fig:policy_prompt}, \ref{fig:processor_prompt}, \ref{fig:query_prompt}, \ref{fig:evidence_prompt}, \ref{fig:verdict_prompt}.

\begin{figure*}[b]
\fboxsep=2pt 
\fboxrule=0.5pt 

\fbox{ \begin{minipage}{\textwidth} %
 Try your best to determine if the given input response is factually accurate. \\

<tool introduction>\\

Use the following format:\\

Response: the response of language model to the user query. you must verify the factual accuracy of the response. If the input is to long, summarize it without changing factualness.\\
Thought: you should always realize what you have known and think about what to do and which tool to use. \\
Action: the action to take, should be one of [actions]\\
Action Input: the input to the action, must follow instructions of tools\\
Observation: the result of the action\\
... (this Thought/Action/Action Input/Observation can repeat N times)\\
Thought: I can give an answer based on the evidence\\
Final Answer: should be in the form: supported, partially supported, not supported, refuted\\

<in-context examples>\\

Begin!\\

<text to verify>
  \end{minipage}}
 
  \caption{Example prompt for the \policyagent{}.}
  \label{fig:policy_prompt}
\end{figure*}

\begin{figure*}
\fboxsep=2pt 
\fboxrule=0.5pt 

\fbox{ \begin{minipage}{\textwidth} %
You and your partners are on a mission to fact-check a claim that may contain multiple subclaims that need to be verified. A sentence that needs to be verified is any statement or assertion that requires evidence or proof to support its accuracy or truthfulness. For example, “Titanic was first released in 1997” necessitates verification of the accuracy of its release date, whereas a claim like "Water is wet" does not warrant verification. Each subclaim is a simple, complete sentence with single point to be verified. Imagine yourself as an expert in processing complex paragraphs and extracting subclaims. Your task is to extract clear, unambiguous subclaims to check from the input paragraph, avoiding vague references like 'he,' 'she,' 'it,' or 'this,' and using complete names.\\

To illustrate the task, here are some examples:\\
<in-context examples>\\

Now, let's return to your task. You are given the following input paragraph, please extract all subclaims that need to be checked.\\

Input: <input>\\
Subclaims: \textcolor{blue}{<extracted claims>}
  \end{minipage}}
 
  \caption{Example prompt for the \claimprocessor{} module. \textcolor{blue}{<Extracted claims>} is the expected output of the LLM for \claimprocessor{}.}
  \label{fig:processor_prompt}
\end{figure*}

\begin{figure*}
\fboxsep=2pt 
\fboxrule=0.5pt 

\fbox{ \begin{minipage}{\textwidth} %
You and your partners are on a mission to fact-check a paragraph. Subclaims requiring verification have been extracted from the paragraph. Imagine yourself as an internet research expert. Your task is to generate a search query for each subclaim to find relevant information for fact-checking. You will be provided with the context of a claim and the specific claim for which you should create a search query.\\

To illustrate the task, here are some examples:\\
<in-context examples>\\

Now, let's return to your task. You are given the following claim and its context, please predict the most appropriate search query for it.\\

Context: <original input text>\\
Claim: <claim to verify>\\
Query: \textcolor{blue}{<predicted search queries>}
  \end{minipage}}
 
  \caption{Example prompt for the \querygenerator{} module. \textcolor{blue}{<Predicted search queries>} is the expected output of the LLM for \querygenerator{}.}
  \label{fig:query_prompt}
\end{figure*}

\begin{figure*}
\fboxsep=2pt 
\fboxrule=0.5pt 

\fbox{ \begin{minipage}{\textwidth} %
You and your partners are on a mission to fact-check a claim. Your mission is to verify a claim's factual accuracy. As experts in reading comprehension, you'll receive a claim and a passage.
You should first read the claim and the passage carefully. Make sure you understand what information you are looking for. Then select sentences that either support, partially support, or refute the claim. A sentence supports the claim if it provides evidence for all statements in the claim. A sentence partially supports the claim if it confirms some details but not all. A sentence refutes the claim if it contradicts any statement in the claim.
Exercise caution in your selection and judgment, avoiding overstatement. Choose the most relevant evidence and refrain from including noisy information. Base decisions solely on provided information without implying additional details.\\

To illustrate the task, here are some examples:\\
<in-context examples>\\

Now, let's focus on your task. You are given a claim and a passage. Please read the passage carefully and copy sentences that contain information supporting or refuting the claim.\\

Claim: <claim to verify>\\
Passage: <passage>\\
Evidence: \textcolor{blue}{<selected evidence>}

  \end{minipage}}
 
  \caption{Example prompt for the \evidenceseeker{}. \textcolor{blue}{<Selected evidence>} is the expected output of the LLM for \evidenceseeker{}.}
  \label{fig:evidence_prompt}
\end{figure*}

\begin{figure*}
\fboxsep=2pt 
\fboxrule=0.5pt 

\fbox{ \begin{minipage}{\textwidth} %
You and your partners are on a mission to fact-check a claim. Your mission is to verify the factual accuracy of a claim using provided evidence. Your partners have collected evidence, and your expertise lies in assessing the claim's factualness based on this evidence.
You are required to determine whether the claim is supported, refuted, or lacks sufficient information based on the provided evidence. The evidence supports the claim if it confirms all statements and details in the claim. The evidence refutes the claim if it contradicts or disproves any statement in the claim. 'Not enough info' applies when the evidence lacks sufficient data, details, or reasoning to support or refute the claim. Even if the evidence supports part of the claim, it should be considered "not enough info" if there is any detail or statement in the claim that cannot be confirmed by the evidence.
Please exercise caution in making judgments and avoid overstatement. Base decisions solely on the provided information without implying additional details.\\

Here are examples to illustrate the task:\\
<in-context examples>\\

Claim: <claim to verify>\\
Evidence: <selected evidence>\\
Analysis: \textcolor{blue}{<verdict prediction>}
  \end{minipage}}
 
  \caption{Example prompt for the \classifier{}. \textcolor{blue}{<Verdict prediction>} is the expected output of the LLM for \classifier{}.}
  \label{fig:verdict_prompt}
\end{figure*}

\section{\dataset{} Dataset}
\label{sec:dataset}
\subsection{Human Annotation Instruction}
\begin{figure*}
    \centering
    \includegraphics[width=\textwidth]{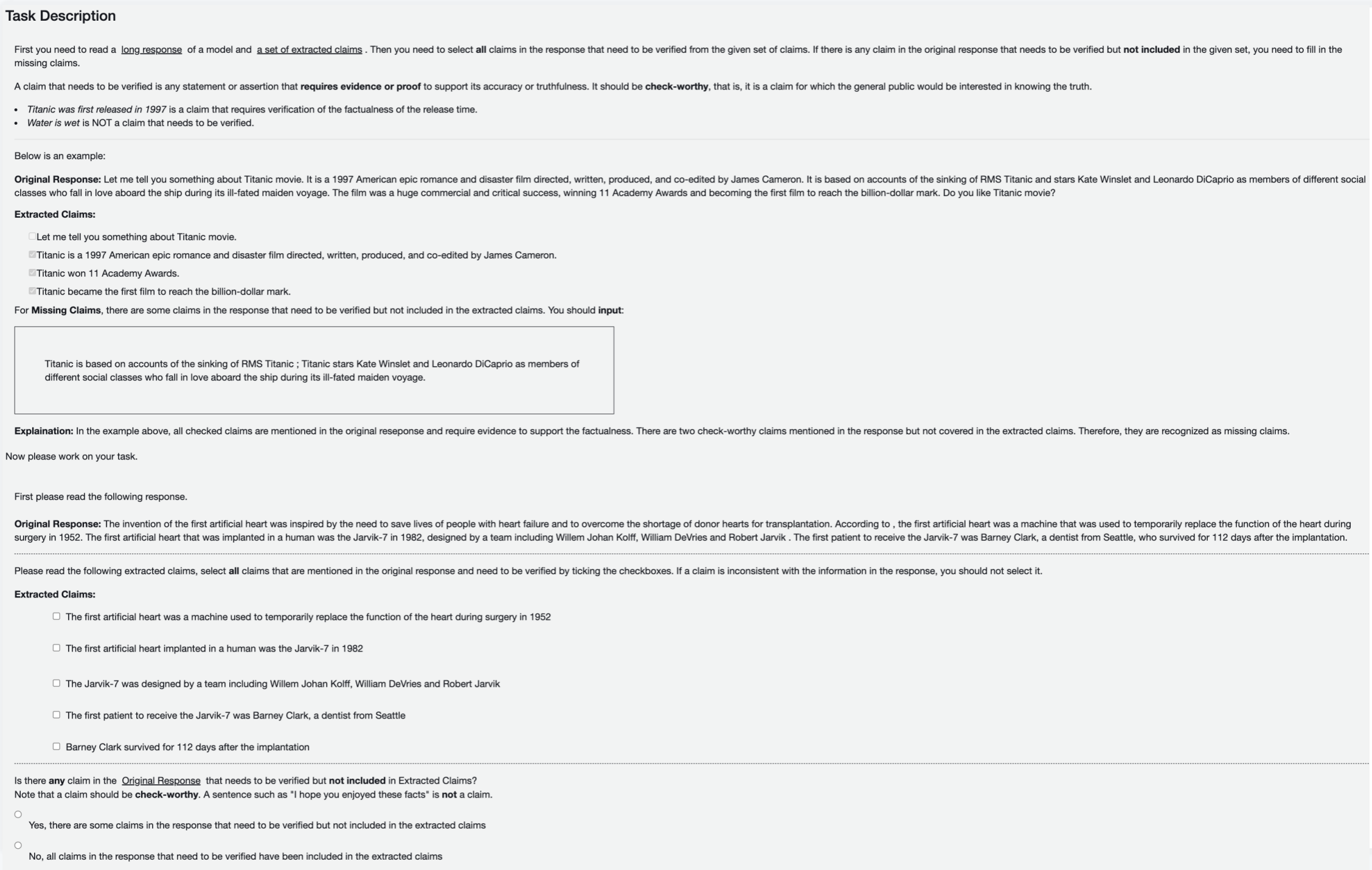}
    \caption{Design of human annotation for claim detection}
    \label{fig:stage1}
\end{figure*}

\begin{figure*}
    \centering
    \includegraphics[width=\textwidth]{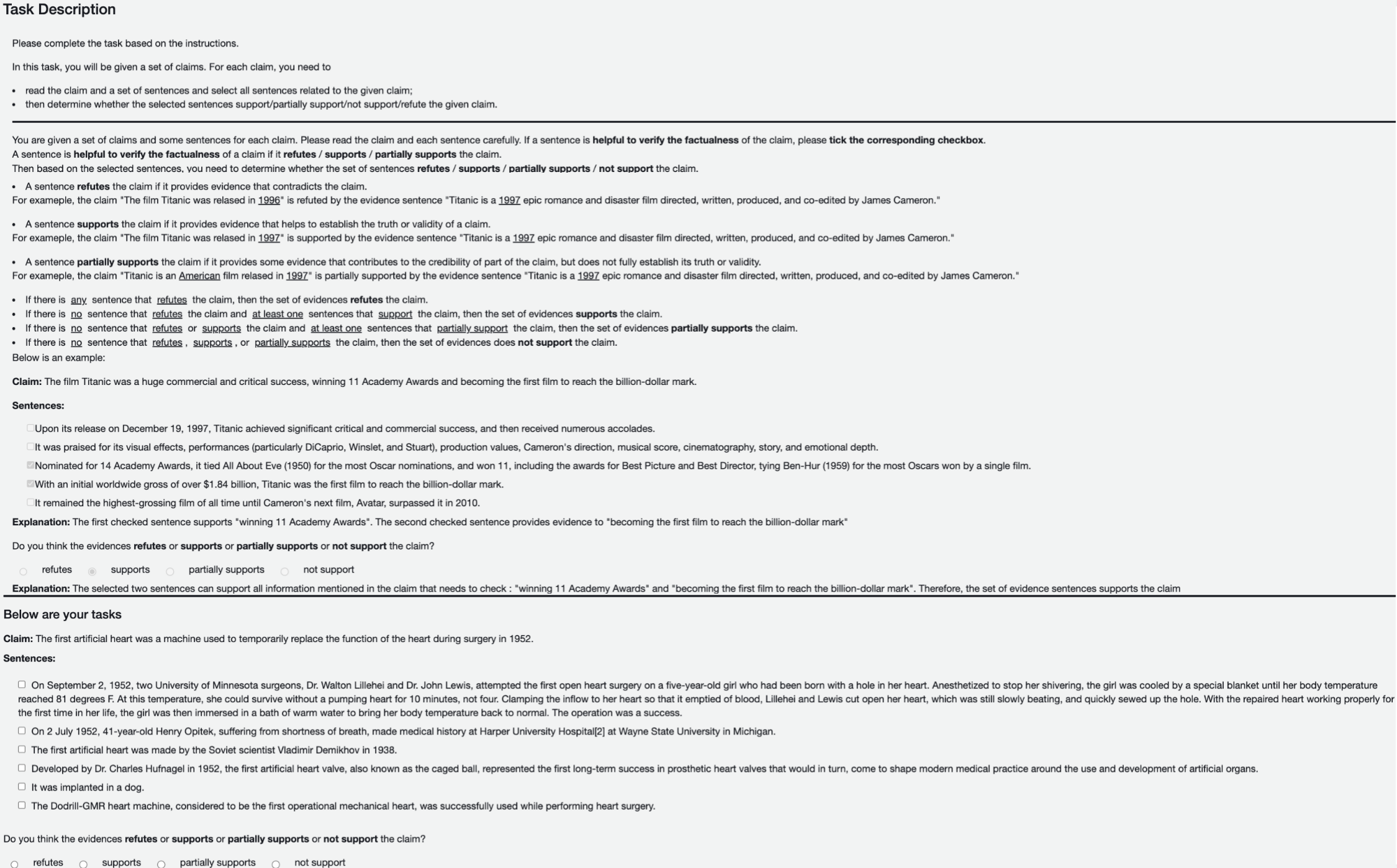}
    \caption{Design of human annotation for evidence retrieval and veracity prediction}
    \label{fig:stage2}
\end{figure*}

We collected human annotated data for \dataset{} in two steps. The design of annotation for claim decomposition is shown in Figure~\ref{fig:stage1}. The design of annotation for evidence retrieval and veracity prediction is shown in Figure~\ref{fig:stage2}.

\subsection{Data Format in \dataset{}}
A record in \dataset{} contains user query, original LLM response, and fact-checking annotation. The fact-checking annotation involves claims to verify, search queries, search results, selected evidence, and verdict labels.
Figure~\ref{fig:dataset_eg} shows an annotated record example.
\begin{figure*}
\fboxsep=2pt 
\fboxrule=0.5pt 

\fbox{ \begin{minipage}{\textwidth} %
\textbf{User Query: } What inspired the invention of the first artificial heart?\\

\textbf{LLM Response: } The invention of the first artificial heart was inspired by the need to save lives of people with heart failure and to overcome the shortage of donor hearts for transplantation. According to, the first artificial heart was a machine that was used to temporarily replace the function of the heart during surgery in 1952. The first artificial heart that was implanted in a human was the Jarvik-7 in 1982, designed by a team including Willem Johan Kolff, William DeVries and Robert Jarvik. The first patient to receive the Jarvik-7 was Barney Clark, a dentist from Seattle, who survived for 112 days after the implantation.\\

\textbf{Fact-checking Annotation: }\\

\begin{minipage}{0.98\textwidth}

\textbf{Claim 1: } The invention of the first artificial heart was inspired by the need to save lives of people with heart failure and to overcome the shortage of donor hearts for transplantation.\\
\textbf{Search Query: } Artificial heart invention inspiration and purpose\\
\textbf{Evidence: }\\
The purpose is to develop a temporary machine or pump for a person who has a disease of the heart and their survival without transplant is impossible.\\
These temporary devices can provide enough time for the patient until a donor heart is available.\\
......\\
\textbf{Label: } Supported\\

\textbf{Claim 2: } The first artificial heart was a machine that was used to temporarily replace the function of the heart during surgery in 1952.\\
\textbf{Search Query: } first artificial heart function replacement machine surgery 1952\\
\textbf{Evidence: }\\
In 1952, the Model II heart-lung device was released to be used on humans.\\
\textbf{Label: } Not Supported\\

\textbf{Claim 3: } The first artificial heart that was implanted in a human was the Jarvik-7 in 1982.\\
\textbf{Search Query: } first artificial heart implanted in human Jarvik-7 1982\\
\textbf{Evidence: }\\
The 61-year-old retired dentist was in an advanced stage of cardiomyopathy, a progressive weakening of the heart muscle, when he became the first recipient of a permanent artificial heart on this day, Dec. 2, in 1982.\\
......\\  
\textbf{Label: } Supported\\

\textbf{Claim 4: } The Jarvik-7 was designed by a team including Willem Johan Kolff, William DeVries, and Robert Jarvik.\\
\textbf{Search Query: } Jarvik-7 artificial heart design team members\\
\textbf{Evidence: }\\
Jarvik completed two years of study, and in 1971 was hired by Willem Johan Kolff, a Dutch-born physician-inventor at the University of Utah,who produced the first dialysis machine, and who was working on other artificial organs, including a heart.\\
......\\   
\textbf{Label: } Partially Supported\\

\end{minipage}
(next page)
  \end{minipage}}
\end{figure*}

\begin{figure*}
\fboxsep=2pt 
\fboxrule=0.5pt 

\fbox{ \begin{minipage}{\textwidth} %
(Continued)\\

\begin{minipage}{0.98\textwidth}
\textbf{Claim 5: } The first patient to receive the Jarvik-7 was Barney Clark, a dentist from Seattle.\\
\textbf{Search Query: } Jarvik-7 first patient Barney Clark Seattle\\
\textbf{Evidence: }\\
On December 2, 1982, Clark became the world\u2019s first recipient of an artificial heart.\\
The 61-year-old retired dentist was in an advanced stage of cardiomyopathy, a progressive weakening of the heart muscle, when he became the first recipient of a permanent artificial heart on this day, Dec. 2, in 1982.\\
......\\
\textbf{Label: } Supported\\

\textbf{Claim 6: } Barney Clark survived for 112 days after the implantation of the Jarvik-7.\\
\textbf{Search Query: } Barney Clark Jarvik-7 implantation survival duration\\
\textbf{Evidence: }\\
Barney Clark survived for 112 days after the implantation of the Jarvik-7.\\
On 1 December 1982, William DeVries implanted the artificial heart into retired dentist Barney Bailey Clark (born 21 January 1921), who survived 112 days with the device, dying on 23 March 1983.\\
\textbf{Label: } Supported\\
\end{minipage}

  \end{minipage}}
 
  \caption{An example in \dataset{}. A record contains a user query, original LLM response, and fact-checking annotation. The fact-checking annotation involves claims to verify, search queries, search results, selected evidence, and verdict labels. The search results and the part of selected evidence are omitted due to space limit.}
  \label{fig:dataset_eg}
\end{figure*}

\clearpage
\section{Ablation Study}
\label{sec:ablation}
\begin{table*}
  \centering
  \resizebox{\textwidth}{!}{
  \begin{tabular}{c|cc|ccccc}
    \toprule
    \multirow{2}{*}{Model} & \multicolumn{2}{c|}{Golden Evidence} & \multirow{2}{*}{FEVER Score} & \multirow{2}{*}{Accuracy} & \multicolumn{3}{c}{Evidence Retrieval}\\
    & Document & Sentence & & & F1 & Precision & Recall \\
    \midrule
    \multirow{3}{*}{\framework{}} & \xmark & \xmark & 51.3 & 62.7 & 55.9 & 64.1 & 49.5\\
    & \cmark & \xmark & 64.1 & 72.8 & 75.8 & 90.3 & 65.4\\
    & & \cmark & - & 81.20 & - & - & -\\
    \bottomrule
  \end{tabular}}
    \caption{Ablation results on entire FEVER test set. ``Golden Evidence'' indicates whether the golden documents/sentences are provided.}
    \label{tab:fever_ab}
\end{table*}
\begin{table*}
  \centering
  \resizebox{\textwidth}{!}{
  \begin{tabular}{c|c|c|cc|ccc}
    \toprule
    \multirow{2}{*}{Model} & \multirow{2}{*}{Golden Evidence} & \multirow{2}{*}{Claim Split} & \multirow{2}{*}{F1} & \multirow{2}{*}{Accuracy} & \multicolumn{3}{c}{Evidence Retrieval}\\
    & & & & & F1 & Precision & Recall \\
    \midrule
    \multirow{2}{*}{T5-3B~\cite{kamoi2023wice}} & \xmark & \cmark & 65.3 & 77.1 & 67.4 & 65.0 & 81.7\\
    &\cmark & \cmark & 78.0 & 84.4 & - & - & -\\
    \midrule
    \multirow{4}{*}{\framework{}}  & \xmark & \cmark & 64.4 & 68.4 & 42.1 & 70.2 & 30.1\\
    &\xmark & \xmark & 35.6 & 35.8 & 23.5 & 91.1 & 13.5\\
    &\cmark & \cmark & 78.7 & 78.8 & - & - & -\\
    &\cmark & \xmark & 71.5 & 47.7 & - & - & -\\
    \bottomrule
  \end{tabular}}
    \caption{Ablation results on WiCE. ``Golden Evidence'' indicates whether golden sentences are provided. ``Claim Split'' indicates whether claim decomposition is performed. Note that we compare with the model finetuned on WiCE dataset~\cite{kamoi2023wice}.}
    \label{tab:wice_ab}
\end{table*}
\paragraph{Ablation Study Results on FEVER dataset} Comparing the first and second rows of Table~\ref{tab:fever_ab}, we observe substantial improvements across all metrics when predicted documents are replaced with golden evidence documents. This improvement suggests the importance of exploring more effective strategies for generating appropriate search queries and improving document retrieval accuracy. Furthermore, the inclusion of golden evidence sentences can further improve the accuracy of veracity prediction by more than 8 points. However, even with golden evidence sentences, the \framework{} lags behind the SOTA model in label accuracy, indicating the need for further enhancements in the \classifier{}'s performance.

\paragraph{Ablation Study Results on Wice Dataset} The evaluation results on WiCE dataset is shown in Table~\ref{tab:wice_ab}. The slight improvement in verdict prediction between the first and third rows of the \framework{} results suggests that the \evidenceseeker{} module's efficiency is unlikely to be the primary bottleneck in the \framework{}'s performance. However, comparing the second row of the baseline with the third row of the \framework{} results highlights that the \classifier{} module's performance is the primary bottleneck in the overall performance of \framework{}. This finding aligns with the results obtained on the FEVER dataset, indicating the significant potential for enhancing verdict prediction despite LLMs' superior capabilities in various NLP tasks. Consistent with prior findings~\cite{kamoi2023wice}, we find that decomposing complex claims into simpler sub-claims improves both evidence retrieval and verdict prediction.

\end{document}